# From Data Topology to a Modular Classifier

Abdellatif ENNAJI, Arnaud RIBERT, Yves LECOURTIER

P.S. I :. Perception, System, Information Laboratory

University of Rouen,

F-76821 Mont Saint Aignan cedex, France

Phone : (33).2.35.14.67.65, Fax : (33).2.35.14.66.18

Contact author : Abdellatif Ennaji

Email : Abdel.Ennaji@univ-rouen.fr

*Author's Biography*

*Abdellatif ENNAJI* has been an Associate Professor at the University of Rouen since 1993. He received a PhD degree from the University of Rouen in 1993 in the field of the cooperation in classification and neural networks for pattern recognition applications. His current research domain concerns problems with Learning, Classification, Data Analysis, and in particular, the problem of data incremental learning of neural networks. These activities are especially applied to pattern recognition problems and decision-making aid in information systems.

*Arnaud RIBERT* received a PhD degree from the University of Rouen in 1998. His major interest was data analysis and neural networks and essentially to a distribution of classification task methodologies. He currently carries out his professional activities in an industrial company.

*Yves LECOURTIER* was born in 1950 in Marseilles. He received a PhD degree in signal processing in 1978, and a second one in Automatic in 1985 from the University of Paris. He was an Associate Professor from 1974 and joined the University of Rouen as a Professor in 1987. His research domain is now in pattern recognition and neural networks. Pr. Lecourtier is a member of AFRIF, ASTI, IAPR. From 1994 to 2000, he was the chairman of the GRCE, a French society which gathers most of the French researchers working in these fields.

**Abstract** :

This article describes an approach to designing a distributed and modular neural classifier. This approach introduces a new hierarchical clustering which enables one to determine reliable regions in the representation space by exploiting supervised information. A Multi-Layer Perceptron is then associated to each of these detected clusters and is charged with recognizing elements of the associated cluster while rejecting all others. The obtained global classifier is constituted of a set of cooperating neural networks and is completed by a k-nearest neighbour classifier which is charged with treating elements rejected by all the neural networks. Experimental results for the handwritten digit recognition problem and comparison with neural and statistical non modular classifiers are given.

# 1. Introduction

Supervised classification task solving in the field of pattern recognition is currently well performed both by neural and statistical algorithms [11]. Neural networks, and more particularly Multi-Layer Perceptrons (MLP) [8,15,33,34] have received a great deal of attention. The reasons for this success essentially come from their universal approximation property [17] and, above all, their good generalisation capabilities, which has been proved for many simple applications in recent years. Comparison of various neural and statistical algorithms have shown that the superiority of one algorithm over another cannot be claimed [11,35]. Performances strongly depend on the characteristics of the problem (number of classes, size of learning set, dimension of the representation space, etc) and on the efforts devoted to the "design phase" of the algorithms (i.e., classifier architecture determination, tuning of learning parameters, etc). Authors in [11] noticed also that a sufficient level of classification accuracy may be reached through a reasonable design effort, and further improvements often requires increasingly expensive design phase [7,11,22]. So, obtaining good generalisation behaviour with an MLP is not a trivial task when dealing with complex problems, since there is no reliable and generic rule currently available to determine a suitable neural network architecture and this can require long trial and error research [2,3,4,6,15].

Moreover, neural networks also have many other defects that are well known and documented [2,3,4,6,10,15,16]. In particular, it has been shown in [13] that the MLP tends to draw open separation surfaces in the input data space, and thus cannot reliably reject patterns. Another drawback of the MLP is the so-called moving target problem: since there is no communication between neurones on a layer, each neurone decide independently which part of the classification problem it will tackle [10].

To overcome these problems, several authors proposed the idea to develop multi-experts decision systems [1,30,37,39,40]. This idea is mainly justified by the need to take into account several sources of information -which can be complementary- in order to reach high classification accuracy and to make the decisions more reliable, and/or to facilitate the classifier design. In this way, several strategies covering most aspects including nature of experts, methods or topologies of decision combination, etc, have been reported in the literature these last years [1,11,30,31,37,39,40].

The approach proposed in this paper redefines the learning task of a neural network so that a simple network building rule can lead to good generalization capabilities with an easy design phase. This redefinition follows a "divide and conquer" strategy with the objective to split the classification problem into several simpler sub-tasks. This classification task distribution is achieved while ensuring a coherency with the data topology in the representation space. The main idea behind this process is to use a "supervised hierarchical clustering" which enables one to determine reliable regions in the representation space. A specialized MLP is then associated to each detected region, and a k-Nearest Neighbour classifier is charged to treat the remaining part of the learning set (non reliable regions). Thus, the whole classifier is a set of cooperative one class neural networks (experts), and it is expected to reach a high accuracy value with simplest learning and designing phases.

Section 2 introduces the basic ideas and general principles used to design a modular and distributed classifier. The method investigated in section 3 achieves the first step of this objective by determining the clusters in the learning set: this clustering provides a « natural » decomposition of the problem. Reliable regions are then obtained through the exploitation of both supervised information (Learning data labels) and unsupervised result of the clustering phase, and will be presented in section 4. The principles of the cooperation scheme into the distributed and modular neural classifier are also presented in section 4. Finally, Section 5 shows experimental results for a handwritten digit recognition problem on the NIST database.

## 2. Distributing a classification problem

Distributing a classification problem presents two main points of interest. The first one is to simplify both the design and the training of a neural network (or any other classifier) by dividing up a given task into several simpler sub-tasks. Such a simplification is expected to lead to an improvement of the generalisation capabilities and accuracy rejection trade-off over those allowed by a single classifier. The second advantage of the classification task distribution is to engineer an easy-to-update modular classifier: when new data are added in the training database, it can be expected that some modules remain unchanged, while some others will just have to be retrained or modified. Moreover, when a new sub-class appears, a new module can easily be added avoiding a complete rebuilt of the classifier.

On the other hand, sub-solutions provided by a modular classifier are usually integrated via a multi-expert decision making strategy. The multi-expert approaches based on experts cooperating differ widely (especially) both with respect to the combination decision strategy [30] and also in the way the problem is approached. In other words, the choice of experts which correspond to the way of splitting the initial task, and/or the context definition of each expert must be considered. The expert context definition include the data representation (set of features) used by each expert, the type of classifier output ...etc. The designer should take into account all these parameters in the combination scheme [11,30,31,36,37,39,40] in order to obtain an optimal behaviour in regards of the classification performances.

In order to obtain high performances without significantly increasing the system complexity, two ways of Multiple Classifier Systems design may be followed. First, one can consider that the chosen set of classifiers providing reasonable accuracy generates a sufficient number of uncorrelated errors. In this case, high accuracy could be reached if an efficient combination strategy is used to exploit this "complementary" behaviour. It must be noticed that such a behaviour is not easy to obtain in the case of a real problem. The second way which can be followed in the design of Multiple Classifier Systems is the above mentioned approach and it consists in decomposing the classification task into simpler sub-tasks, each one being handled by a separate expert. This approach differs from the first one mainly in the sense that no potential and explicit "complementary behaviour" between experts is expected, the main goal of such a method being to specialise each expert in a particular sub-task. The task decomposition should produce sub-tasks as simple as possible allowing a simple and robust classification modules. The multi-experts decision-making module, even simpler, usually has enough information to make an accurate global decision.

In this paper, a cooperative modular Neural Network, in combination with a statistical classifier, is introduced with the objective to improve the performances in terms of robustness, adaptability and accuracy-rejection trade-off over those allowed by a single classifier. In light of the above, the proposed approach tends to reach this objective by splitting the initial classification task into a simpler sub-tasks obtained by extracting the topology of the learning data set in the feature space.

Thus, we propose an unsupervised procedure which provides an automated task decomposition without any *a priori* knowledge. The simplest problem to be given to a neural network is probably a linearly separable one. Unfortunately, few real problems are as simple as this. A second type of simple classification problem - although more complex than the previous one - may be encountered for a two class problem such that at least one of them is constituted by an homogeneous region (i.e. a unique pure cluster containing elements of the same class only). The proposed "divide and conquer" strategy aims to identify this kind of cluster - called an "islet" - in order to provide as many tasks as islets. If N islets are detected, the classifier will thus be constituted from N cooperative neural networks, each of them being quite simple to configure and being expected to present good decision boundaries. In other words, the partial goal of this approach remains to obtain a classifier capable of defining a closed separation surfaces in the feature space allowing a reliable rejection behaviour. Such a behaviour is required in applications like pattern recognition [13].

This approach can be compared with Jacobs and Jordan's work [18]. However, their approach is based upon a completely supervised learning. Consequently, the number of experts has to be defined by the user. In the same way, this approach is close to the work presented in [16] where an unsupervised self-organised network is used to clusterize a MLP in order to avoid the "moving target" problem. This work does not allow any cooperation scheme nor a modular system but seems able to perform an incremental learning. So, the most important and common limitation of these works is to require to know the number of clusters in learning data which is not easy when dealing with a real and complex problem. Note that designing a distribution scheme according only to the supervised information is not always appropriate in regards with the objective to make decisions reliable, it seems intuitive and important to take into account the real distribution of the learning data in the feature space.

## 3. Multi-Level Hierarchical Clustering

As mentioned above, the first stage towards the classification task distribution consists in capturing the data structure in the feature space. To achieve this, the problem decomposition starts with a clustering phase in order to extract reliable clusters or regions. Thus, it implies to determine the number and the constitution of the clusters in the learning data set. The most commonly used techniques are certainly Self-Organizing Maps [23] and partitional clustering methods (like k-means [25]). The term « partitional clustering » is used in contrast to hierarchical clustering, where hierarchical is in fact based on a nested sequence of partitions (in the same way as Jain & Dubes [19]). The main problem with these methods is that in practice, the number of clusters is required in advance to obtain a good representation of the data. Moreover, in the case of partitional methods, the usually used quadratic criterion of cluster compactness leads to hyper-spherical groups, which does not necessarily match the reality. Consequently, when clusters do not exhibit compact spherical shapes, partitional methods often provide a low quality clustering. However, well known methods like Hierarchical clustering have been known for long time, which explore the hierarchical structure of a feature space. As the first step of our approach, we will introduce a new algorithm of cluster detection from a hierarchical tree.

### 3.1- Hierarchical Clustering

Hierarchical clustering methods give a graphical data representation without any assumption on *a priori* distribution nor on the number of clusters. An example of the corresponding graphical representation, called a dendrogram, is shown on Fig. 1. It is important to note that the height of a node represents the distance between the groups it links. This explains why the shape of a dendrogram gives information on the number of clusters in a data-set. Hierarchical clustering algorithms generally proceed by sequential agglomerations of clusters. A hierarchical clustering can be built using the following algorithm, where initial points are considered as individual clusters [19].

---

Compute the Euclidean distance between every pair of points;

Merge into a single cluster the two closest points;

**While** ( There are more than one group ) **Do**

       Compute the distance between the new cluster and every existing one;

       Merge into a single cluster the two closest ones;

**EndWhile**

---

Fig.1

It can be noticed that the Euclidean distance can be replaced by any other dissimilarity function. Moreover, an additional metric has to be introduced to measure the distance between two clusters $C_1$ and $C_2$. One of the most commonly used metrics, called single link algorithm, consists of computing the minimum distance between two points $X_1 \in C_1$ and $X_2 \in C_2$. The maximum and average links are also often encountered but an infinite number of metrics can be obtained using Lance-Williams' formulae [24] and its generalisation [20]. It is therefore possible to adapt the metric to a particular problem. This choice has a great influence on the representation capabilities of the hierarchy, and then, on the possibilities of clusters extraction. Indeed, authors in [36] introduce a new technique of agglomerative hierarchical clustering in order to avoid this problem and introduce a criterion of cluster merging which makes it possible to generate n-ary hierarchies more adapted and easy to interpret for certain problems.

In order to optimise the building process, and to work with a well-suited metric, the Lance-Williams' formula has been used here, so that the resulting metric is close to the single link, but avoids the associated chaining effect (data tend to be merged into a single cluster). It is then possible to obtain large clusters whose shape is not necessarily hyper-spherical [32].

### 3.2- Clusters Detection

Once the hierarchy is built, commonly used methods to determine a clustering from a dendrogram consists of cutting it horizontally. As stated by Milligan and Cooper [26], numerous methods have been proposed to find the best cutting point, most of them use statistical criterions such as the maximisation/minimisation of the variance inter/intra-clusters. Although most of them will detect 5 clusters in the (simple) example of Fig. 1, unique cutting point methods cannot efficiently treat real data in the general case.

Indeed, when dealing with a great amount of real data, the built dendrograms are rarely as easy to interpret as the previous one. One of the problems that can arise is a large variation of the node heights due to important variations of the density in the data. Fig. 2 gives a synthetic example of such a configuration which cannot be efficiently treated by a unique cutting in the dendrogram. Roughly, three situations may occur. Most of the time, the clustering method will detect four clusters, thus ignoring the structure of the densest groups. It can also be considered that the number of clusters is known (although this is rarely verified in practice). At this time, a unique cutting in the dendrogram will not lead to detect the three dense clusters, but will split the three least dense clusters. Finally, to detect the three densest clusters by a unique cutting, 17 clusters will appear, which is far from reality. This phenomena appears also on the real problem (Handwritten digit recognition) as shown in [32]. Consequently, it can be said that classical methods should only be used when the data density is almost constant in the representation space.

Fig. 2

The problem of density variations leads to introduce several cutting points in a dendrogram. The configuration in Fig.2 shows that (i) the data have first to be considered from a global point of view and (ii) that certain dense (and well structured) parts have to be detailed to obtain a reliable representation of the reality. This is why a Multi-level Hierarchical Clustering [32] is proposed in this paper. It is based upon the detection of a single cluster on a given sub-tree. Thus, if during the descendant exploration, the current sub-tree corresponds to a single cluster, no further investigation is necessary in this sub-tree. On the contrary, the exploration goes on.

Identifying a single cluster in a sub-tree is known in the literature as a "cluster validation" problem. Different bibliographical studies [12,19,20] show that most of the techniques require a strong assumption upon the statistical distribution of the expected clusters. This condition being intractable in the general case, we propose an original technique. Moreover, in order to have a fast procedure, all the allowed computations exclusively involve the hierarchy itself. Consequently, no direct use of original data will occur.

From these requirements, one needs a criterion to measure the shape of a dendrogram, i.e. the distribution of the nodes inside the tree. This is done by computing an histogram of the heights of the nodes. A typical histogram for a binary hierarchy representing a single cluster appears in bold lines in Fig. 3.b. If the treated data formed two clusters, the main changing in the hierarchy would concern its top (in dash lines in Fig. 3.a), which would be higher than for a single cluster. Fig. 3.a illustrates an ideal situation where all the levels but the top are equal (this assumption is only made in a simplification purpose, without any loss of generality). Then, it can be stated that ($Level_{Max}^{2Cl} - Level_{Min}^{2Cl}$) is superior to ($Level_{Max}^{1Cl} - Level_{Min}^{1Cl}$). Consequently, the histogram bin-widths are wider for $H_1$ (1 cluster) than for $H_2$ (2 clusters). Since the levels of both hierarchies are equal, except for the highest one, it can be said that the very first bars of $H_2$ will be taller than their equivalents in $H_1$. Conversely, some of $H_2$ intervals will contain no element (see Fig. 3.b). At last, $H_2$ will present a last interval with 1 element (the top of the corresponding hierarchy). As a conclusion, it can be said that an histogram is sensitive to the number of clusters in the data set. This sensitivity has been measured by the standard deviation of the tallness of an histogram bars, s, divided by their average, m. This criterion, called *variation coefficient* in the literature [38] is employed as a measure of the homogeneity of a population. It can be noticed that in our case, this criterion is particularly interesting since it increases with the number of clusters in a sub-tree.

Fig. 3

This measure is employed in an algorithm whose principle is to explore the hierarchy in depth in order to detect a relevant changing in the number of clusters. At a given node, the method consists of assuming an horizontal cutting in the hierarchy (following a simple technique which is detailed further). This cutting reveals a certain number of sub-trees. This assumption is validated if the number of clusters detected in each sub-tree is not too high in comparison to the number of clusters in the tree. Indeed, theoretically, when performing an in depth exploration, the number of clusters in a sub-tree has to be inferior to the number of clusters represented by its father. Consequently, if the detected number of clusters increases, it can be stated that the cutting was abusive. When dealing with real data, this principle is not always strictly respected, and a tolerance margin has to be introduced.

Globally, it can be said that the method described in this paper replaces a unique horizontal cutting by several well chosen ones. The horizontal cutting method that has been employed consists in sorting the values of the hierarchy and then in computing the highest difference between two consecutive values ($V_{n+1}$ - $V_n$). The cutting is then performed just below the level n+1. This method has been preferred to a simple dichotomic exploration because at the end of the process, it leads to systematically decide whether there is one or two clusters in a given sub-tree, which is the most difficult for our criterion (which can more easily detect the presence of multiple clusters). Eventually, the employed algorithm is described below:

---

**Begin**

    CurrentFatherNode = Root of the hierarchy;

    Make the assumption of a unique horizontal cutting in the tree of root CurrentFatherNode;

    **For** Each discovered sub-tree **Do**

        CurrentNode = Root of the Current Sub-Tree;

        **If** ( s/m(CurrentNode) > α.s/m(CurrentFatherNode) ) **Then**

            Invalidate the horiz. cutting : elts of CurrentFatherNode belong to the same cluster;
            Stop the recursive exploration of CurrentFatherNode;

        **Else**

            CurrentFatherNode = CurrentNode;

            Recursively explore the tree of root CurrentFatherNode;

        **End If**

    **End For**

**End**

---

As can be seen, this algorithm only requires the determination of a single parameter, α,which defines the suitable tolerance in order to deal with noisy data. This process is carried out using a small fraction of the database as a learning set. Since the number of clusters increases with α, a dichotomic research is particularly fast and efficient: α is initiated with a high value (e.g. 10), and quickly converges towards a suitable value. The function to be optimised may be given by any statistic criterion. An example of such function is given in [32] and allow us to evaluate the clustering quality in the supervised meaning. More details and discussions of this method are also given in [32], and Fig.4 gives the obtained result of the example of Fig.2 without any parameter estimation ($\alpha$ is fixed arbitrarily to 1).

## 4. Construction of Cooperative Modular Classifier

The second step of the modular and cooperative classifier design consists in exploiting the clustering result in order to extract the data topology. This extraction is made by labelling the vectors according to their class in the supervised meaning. In this way, it is possible to compare the supervised and unsupervised information provided respectively by a vector label and the dendrogram. Afterwards, an analysis of the composition of the sub-trees reveals the presence of islets (i.e. clusters comprising at least P elements from the same class, P being user-defined and correspond to the minimum allowed size of an islet). Fig. 5 illustrates the resulting distribution after applying such a technique.

Each islet is then learnt by an MLP which has to solve a two class problem: recognize its associated islet while rejecting all other elements. Since the elements of an islet are close to one another, it can be expected that the problem is simple enough to allow a basic MLP building rule to be efficient.

Another appropriate solution could be the use of Radial Basis Functions (RBF) networks [27,28] rather than MLP for cluster identification. Because of their local learning properties, the RBF-like networks seem *a priori* better adapted to achieve this task. However, they also have significant defects which make them less attractive. In particular, they usually require a very large number of neurones in order to learn a smooth transformation between input-output space [15] (mainly in the case of high dimension input space), and more training data are required to achieve similar precision to that of the MLP network [14]. In spite of that, it seems an interesting way to explore in future work.

As might be expected, all elements will not be associated with some islet. Indeed, some of the elements are located near the boundaries, so they will not appear in a pure sub-tree. Moreover, depending on P, the minimum allowed size of an islet, large fluctuations on their number may arise. The percentage of the elements of the learning database associated with an islet may therefore vary in a large scale too. Consequently, the neural networks do not learn the whole database but only its most reliable parts. A supplementary classifier has thus been used to obtain high recognition performances. Non-parametric methods seem to be best-suited, since they do not require any real learning stage and their computing cost can be widely reduced by the cooperation process. Furthermore, this remaining part of the learning set seems to belong to non reliable regions (non homogenous clusters, overlap regions...). In practice, a K-nearest neighbour (K-NN) classifier has been used for this task.

Cooperation with neural networks has been implemented as follows : apply all the neural networks for the unknown pattern; if only one of them recognises the element, take this decision, otherwise take the decision of the K-NN. This combination strategy is basically simple knowing that many other and often more sophisticated methods are available [30,31,37,39,40]. Indeed, such an option was made in order to emphasize the consequent influence in the performances improvement and in the simplification design phase by the fact that the data topology was taken into account in the distribution task. Other more sophisticated and complex combination methods could be used and there remains opportunities for future developments and improvements of this work.

This hierarchical architecture of the neural classifier is, in its principle, close to a decision tree [9,29]. In fact, it can be seen as a particular decision tree where the nodes and leaves are neural networks. This architecture can also been compared with nested neural networks [21]. The main difference with this last, resides in the fact that in our case, the task assignation of the different modules (neural networks) is based upon a cluster analysis.

## 5. Experimental results : a handwritten digit recognition problem

Tests have been performed for a handwritten digit recognition problem over the NIST database 3. The feature vector considered is constituted by the 85 (1+4+16+64) grey levels of a 4 level resolution pyramid [5]. Fig.6a gives an example of digits from NIST database when Fig.6b shows an example of the retained representation.

Fig. 6a        Fig. 6b

In order to test the generalisation capabilities of the distributed neural classifier, it has been compared to two reference classifiers : a classical K nearest neighbours (KNN) and a single MLP whose architecture and parameters have been determined by a trial and error procedure. Training and test databases consisted of respectively 20,000 and 61,000 elements. Neural networks involved in the distributed classifier were trained using the classical back-propagation algorithm while their structure was found applying a simple rule : several architectures were considered (comprising 2, 5, 10, 15, 20, 25, 30, 35, 40, 45, 50, 100 hidden units, and 2 hidden layers of 50-20 units) ; when the speed of convergence of a given architecture was too low, the next configuration was tested. The first architecture to achieve success in learning all the elements was retained.

The curves in Fig.7 represent the average error rate (i.e. Number of bad decisions / Total Number of presented digits) according to the recognition (i.e. Number of recognized digits/Total Number of presented digits), over 5 different training and test databases following a cross-validation procedure. Rejection rate is thus given by 100-(error+recognition). These curves have been obtained following two different ways, depending on the considered classifier. The K-NN curve is obtained in decreasing k, while requiring that the k nearest neighbours are of the same class (the maximum recognition rate is obtained when k = 1). The neural network curves are obtained in increasing the minimum value of the maximum output of the network: high thresholds will generate low error rates.

An average of 120 islets of more than P=15 elements were detected, and 76% of the training set was assigned to an islet. Most of the neural nets (88%) presented a single layer of 10 hidden units, while two had 2 hidden layers (of 50-20 units). It can thus be said that, as expected, learning an islet is a rather simple problem.

These statistics show that the modular classifier always present better performances than the single MLP. The performance curve of the distributed classifier is close to the K-NN one, except for low error-rates. Thus, for a 0% error rate, the recognition rate of the distributed classifier is 14% higher than the K-NN one, whereas in this configuration only 41% of the decisions are taken by the networks. This difference shows a fundamental dissimilarity between built boundaries (which are implicit in the case of the K-NN algorithm). Training neural networks on islets enables them to recognize an element from a given class, whereas its 50 or 55 nearest neighbours are from another class. Boundaries generated by the learning of islets (even following a simple network building rule) are therefore particularly efficient. It can be noticed that single neural networks rarely implement such boundaries since no explicit learning rule exist to find them.

## 6. Conclusion

This article deals with the problem of finding a well-suited modular classifier architecture for a given problem. The proposed solution consists in splitting the classification problem in several simpler sub-problems which are determined by a supervised hierarchical clustering procedure. The first experimental results for a real classification task are promising, since they show that training a neural network to solve such a sub-problem leads it to define efficient decision boundaries, specially when low error rates are required. Indeed, a simple network building strategy permitted the recognition of difficult patterns for the K-NN classifier, and produced better results than a purpose-designed MLP. The reliability of neural network decisions for low error rates is thus significantly improved. Further experiments should lead to a better characterisation of these boundaries to provide explicit rules for high-performance network building.

List of figures and tables

Fig. 1 : An example of hierarchical clustering

Fig. 2 : Dealing with high density differences with a classical hierarchical clustering

Fig. 3 : Histograms for 1 and 2 clusters

Fig. 4 : A Multi-level Hierarchical Clustering

Fig. 5 : An example of determination of islets and meta-islets

Fig. 6a: Examples of digits from NIST database

Fig. 6b: Representation of a digit 2 using a 4 level resolution pyramid

Fig. 7 : Performances of 3 classifiers : 20,000 instances on training set; 61,000 on test set

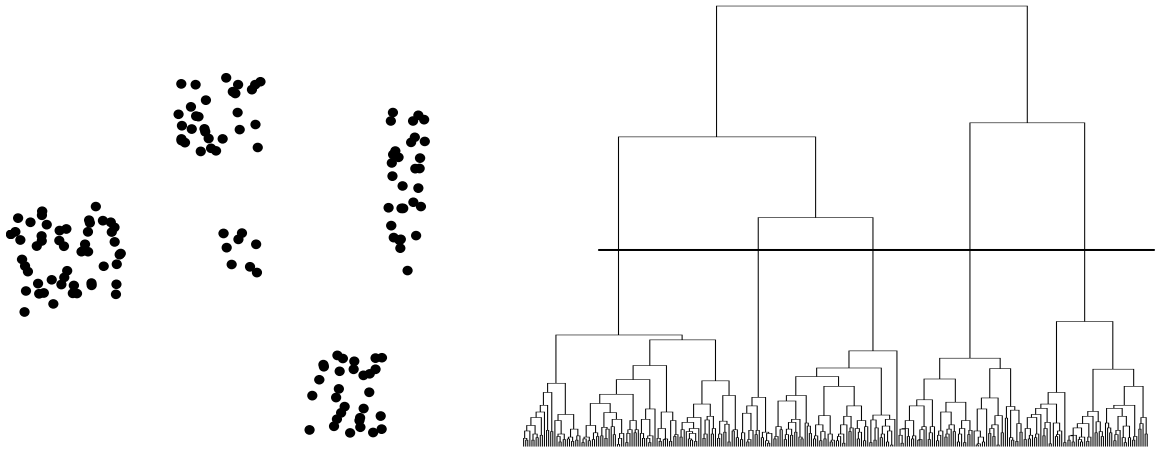

Fig. 1

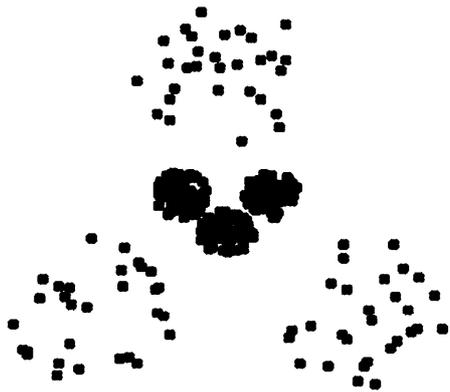

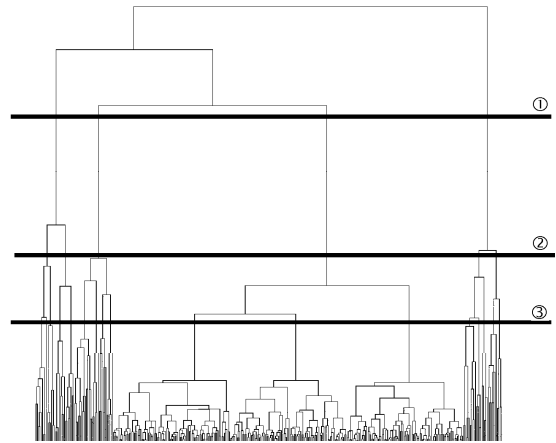

High density variations in the representation space

① Four clusters detected
② Six clusters detected
③ The three dense clusters are detected

Fig. 2

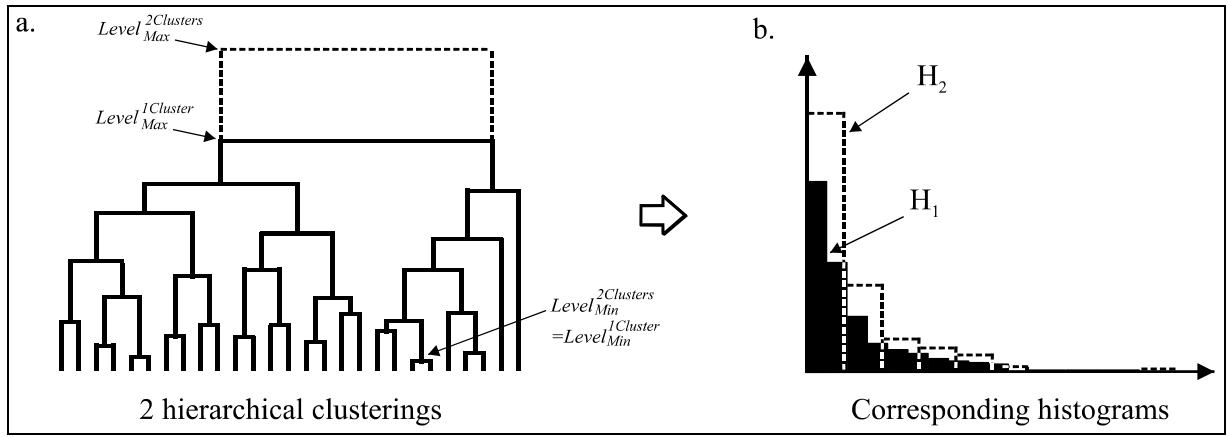

2 hierarchical clusterings          Corresponding histograms

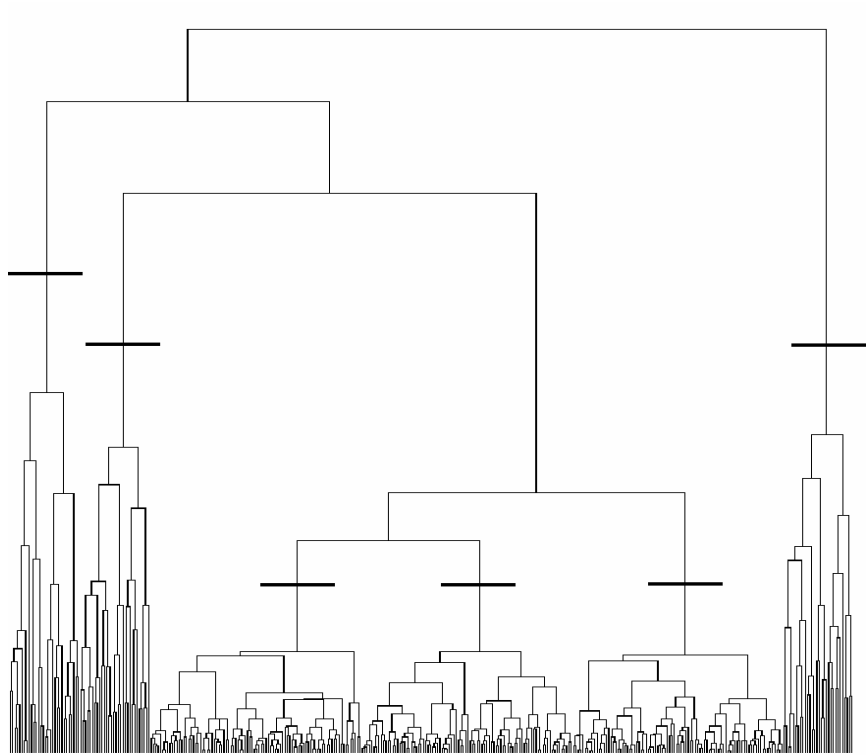

Fig. 4

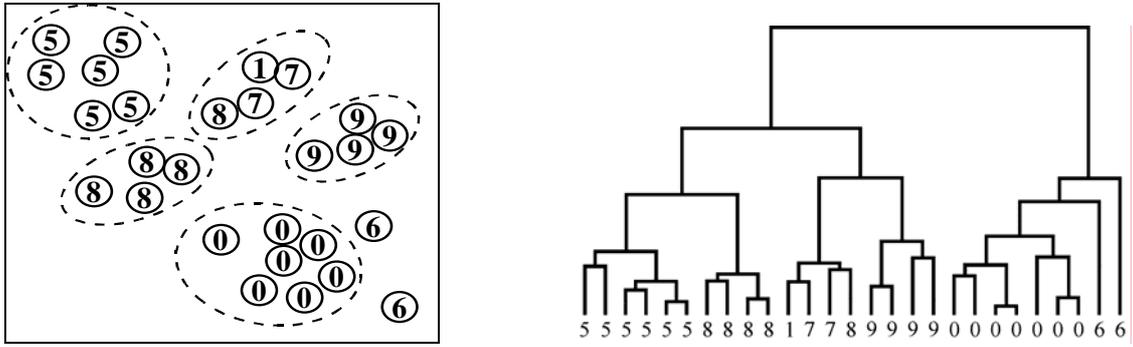

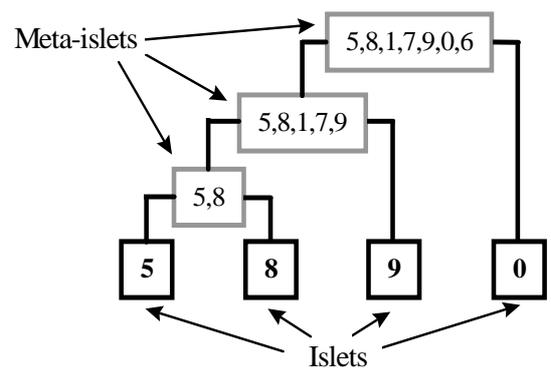

Fig. 5

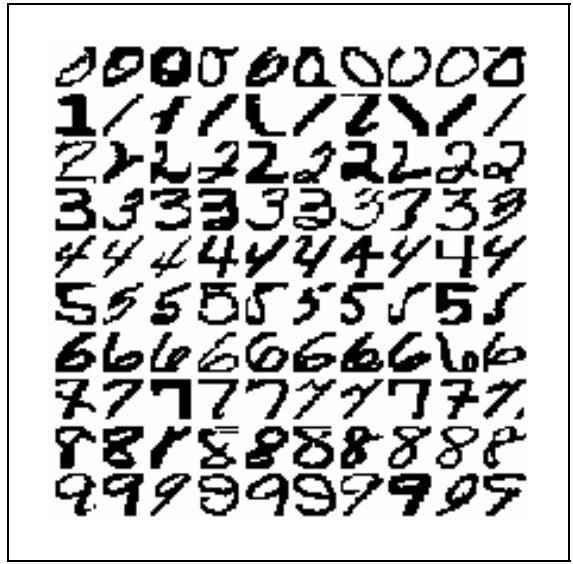

Fig. 6a

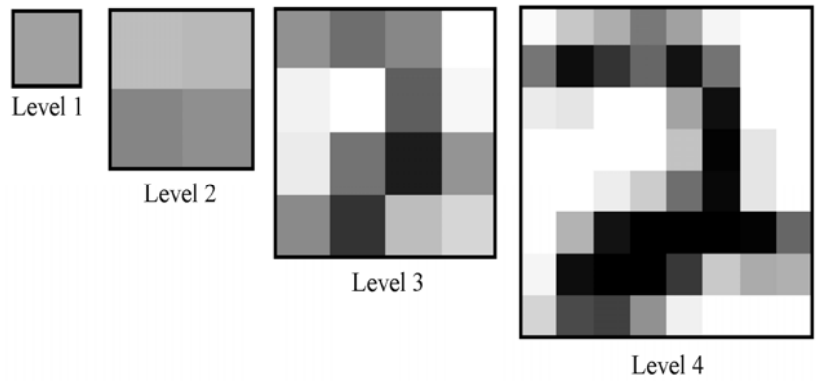

Level 1

Level 2

Level 3

Level 4

Fig. 6b

| Error | Ref. K-NN | Ref. MLP *50-20 h.u.* | Distributed classifier |
|-------|-----------|------------------------|------------------------|
| Max.  | 98.8      | 97.9                   | 98.8                   |
| 0.5%  | 97.5      | 93.5                   | 97.6                   |
| 0.4%  | 96.9      | 92.3                   | 97.0                   |
| 0.3%  | 96.0      | 90.1                   | 96.1                   |
| 0.2%  | 94.1      | 85.5                   | 94.3                   |
| 0.1%  | 89.9      | 69.0                   | 90.6                   |
| 0.0%  | 41.0      | 15.1                   | 54.8                   |

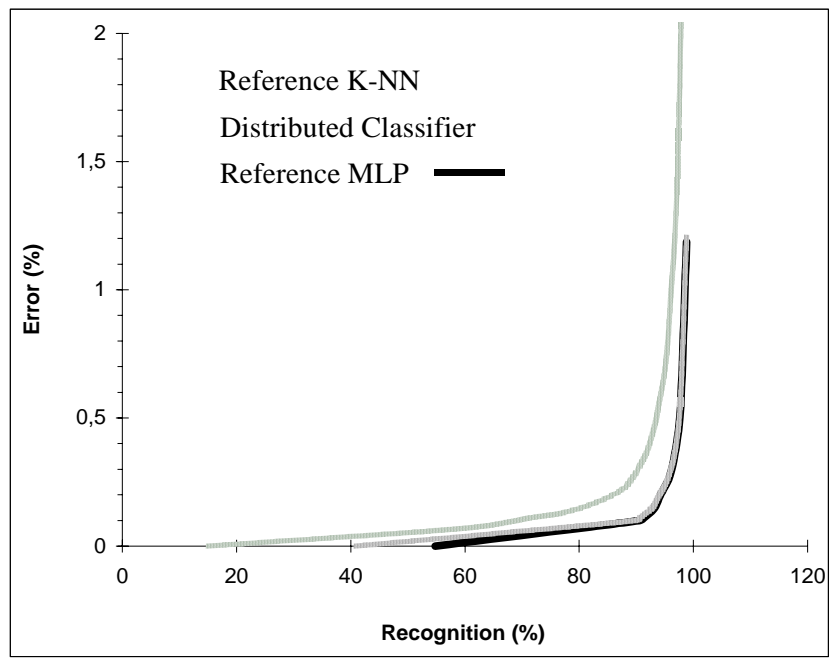

Fig. 7